\documentclass[conference]{IEEEtran}

\usepackage{amsmath}
\usepackage{graphicx}
\usepackage{cite}
\usepackage{algorithm}
\usepackage{algpseudocode}
\usepackage{hyperref}

\begin{document}

\title{Neural Network-Based Tracking and 3D Reconstruction of Baseball Pitch Trajectories from Single-View 2D Video}

\author{
    \IEEEauthorblockN{Jhen Hsieh}
    \IEEEauthorblockA{Department of Computer Science \& Information Engineering\\
    National Taiwan Univeristy \\
    Email: spen940408@gmail.com}
}

\maketitle

\begin{abstract}
In this paper, we present a neural network-based approach for tracking and reconstructing the trajectories of baseball pitches from 2D video footage to 3D coordinates. We utilize OpenCV's CSRT algorithm to accurately track the baseball and fixed reference points in 2D video frames. These tracked pixel coordinates are then used as input features for our neural network model, which comprises multiple fully connected layers to map the 2D coordinates to 3D space. The model is trained on a dataset of labeled trajectories using a mean squared error loss function and the Adam optimizer, optimizing the network to minimize prediction errors. Our experimental results demonstrate that this approach achieves high accuracy in reconstructing 3D trajectories from 2D inputs. This method shows great potential for applications in sports analysis, coaching, and enhancing the accuracy of trajectory predictions in various sports.

\end{abstract}


\section{Introduction}
Tracking and reconstructing the trajectories of baseball pitches is a challenging task due to the high speed and complex dynamics of the ball's motion. Traditional methods, such as the MLB Statcast and Hawk-Eye systems, rely on multiple high-speed cameras and sophisticated calibration techniques to achieve accurate tracking and reconstruction. While these systems are highly accurate, they are also expensive and require extensive setup and maintenance.

In contrast, we propose a neural network-based approach that requires only a single 2D video input to accurately reconstruct the 3D trajectory of a baseball pitch. Our method leverages OpenCV's CSRT algorithm to track the baseball and fixed reference points in 2D video frames. These tracked pixel coordinates serve as input features for a custom neural network model designed to map 2D coordinates to 3D space. By training the model on a dataset of labeled trajectories, we achieve high accuracy in predicting 3D trajectories from single-view inputs.

This approach significantly reduces the hardware requirements and setup complexity, making advanced pitch tracking technology more accessible and affordable. We detail our methodology, including the tracking algorithm and neural network architecture, and present experimental results demonstrating the efficacy of our approach.

\section{Related Work}
The problem of tracking and reconstructing the trajectories of fast-moving objects like baseballs has been extensively studied. Traditional methods, such as the MLB Statcast and Hawk-Eye systems, employ multiple high-speed cameras and sophisticated calibration techniques to achieve high accuracy in tracking and reconstructing the trajectories of baseball pitches. The MLB Statcast system, for instance, uses an array of 12 cameras to capture the motion of the ball from various angles, providing detailed data on pitch trajectory, spin rate, and velocity. While these systems are highly accurate, they are also expensive and require extensive setup and maintenance.

In recent years, computer vision techniques have been increasingly applied to the problem of motion tracking. Monocular and stereo vision systems have been used to track the motion of objects in 3D space. For example, methods based on feature matching and triangulation have shown promise in reconstructing trajectories from 2D images. However, these methods often require precise camera calibration and can be computationally intensive.

Deep learning approaches, particularly those utilizing neural networks, have also been explored for object tracking and 3D reconstruction. Convolutional neural networks (CNNs) and recurrent neural networks (RNNs) have been employed to learn and predict the motion patterns of objects from video data. These methods leverage the powerful feature extraction capabilities of neural networks to achieve high accuracy in tracking and reconstruction tasks. For instance, several studies have demonstrated the effectiveness of using CNNs for end-to-end learning of object trajectories from video frames.

Our proposed method distinguishes itself from these existing approaches by utilizing a neural network-based model that only requires a single 2D video input to accurately reconstruct the 3D trajectory of a baseball pitch. By leveraging OpenCV's CSRT algorithm to track the baseball and fixed reference points in 2D video frames, our approach reduces the hardware requirements and setup complexity. Additionally, our method incorporates timestamp information to compensate for the lack of depth information in single-view setups, providing a cost-effective and accessible alternative for tracking and analyzing baseball pitch trajectories .

In summary, while traditional multi-camera systems offer high accuracy, they are often impractical for many applications due to their cost and complexity. Our neural network-based approach provides a simpler and more affordable solution, with promising results demonstrated in our experimental evaluations.

\section{Methodology}

\subsection{Data Generation}
\begin{itemize}
    \item \textbf{Simulation of Ball Trajectories}: Generate different trajectories of balls to simulate real-world scenarios.
    \item \textbf{Projection onto Screen}: Project the simulated trajectories along with five reference points onto a screen to obtain pixel coordinates.
    \item \textbf{Timestamp Addition}: Include timestamps for each data point to capture the temporal aspect of the trajectories.
    \item \textbf{Input-Output Format}: Define the input format as pixel coordinates along with timestamps and output as the position of the ball in the 3D scene.
\end{itemize}

\subsection{Model Architecture}
\begin{itemize}
    \item \textbf{Neural Network Structure}: Utilize a neural network architecture for training, with the following layers:
    \begin{itemize}
        \item Input layer: Accepts pixel coordinates and timestamps.
        \item Hidden layers: Comprising multiple fully connected layers (fc1 to fc7) with varying hidden sizes.
        \item Output layer: Outputs the position of the ball in the 3D scene.
    \end{itemize}
\end{itemize}

\subsection{Training Process}
The training process involves the following key components:

\begin{itemize}
    \item \textbf{Loss Function}: Define an appropriate loss function to measure the discrepancy between predicted and actual 3D positions.
    
    \item \textbf{Optimizer}: Select an optimizer (e.g., Adam) to update the neural network's parameters during training.
    
    \item \textbf{Training Procedure}:
    \begin{itemize}
        \item Split the dataset into training and validation sets.
        \item Iterate through each epoch of training.
        \item Forward pass: Compute the predicted outputs using the input data.
        \item Loss computation: Calculate the loss between the predicted outputs and the ground truth labels.
        \item Backward pass: Compute the gradients of the loss with respect to the model parameters.
        \item Update model parameters: Adjust the model parameters using the optimizer based on the computed gradients.
        \item Monitor performance: Optionally, validate the model's performance on the validation set to monitor for overfitting.
    \end{itemize}
\end{itemize}

This training process ensures that the model learns to accurately predict the 3D positions of baseball pitches by minimizing the defined loss function through gradient-based optimization.

\subsection{Evaluation Metrics}
\begin{itemize}
    \item \textbf{Accuracy}: Measure the accuracy of the model in predicting the 3D position of the ball.
    \item \textbf{Loss}: Track the loss function during training to ensure convergence and assess model performance.
    \item \textbf{Squared Distance}: Calculate the squared distance between the predicted and actual locations of the ball in the 3D scene. This metric quantifies the discrepancy between predicted and ground truth positions.
\end{itemize}

\subsection{Summary}
The methodology involves generating synthetic data representing ball trajectories, training a neural network model using the generated data, and evaluating the model's performance based on predefined metrics. The neural network architecture consists of multiple fully connected layers, and training is conducted using an appropriate optimizer while monitoring loss and accuracy metrics.

\subsection{Neural Network Architecture}
Detail the architecture of the neural network, including the convolutional layers used for feature extraction and the recurrent layers for sequence prediction.

\subsection{Training Process}
Explain the training process, including the loss function, optimization algorithm, and any data augmentation techniques employed.

\subsection{Results}

\begin{figure}[htbp]
    \centering
    \includegraphics[width=0.4\textwidth]{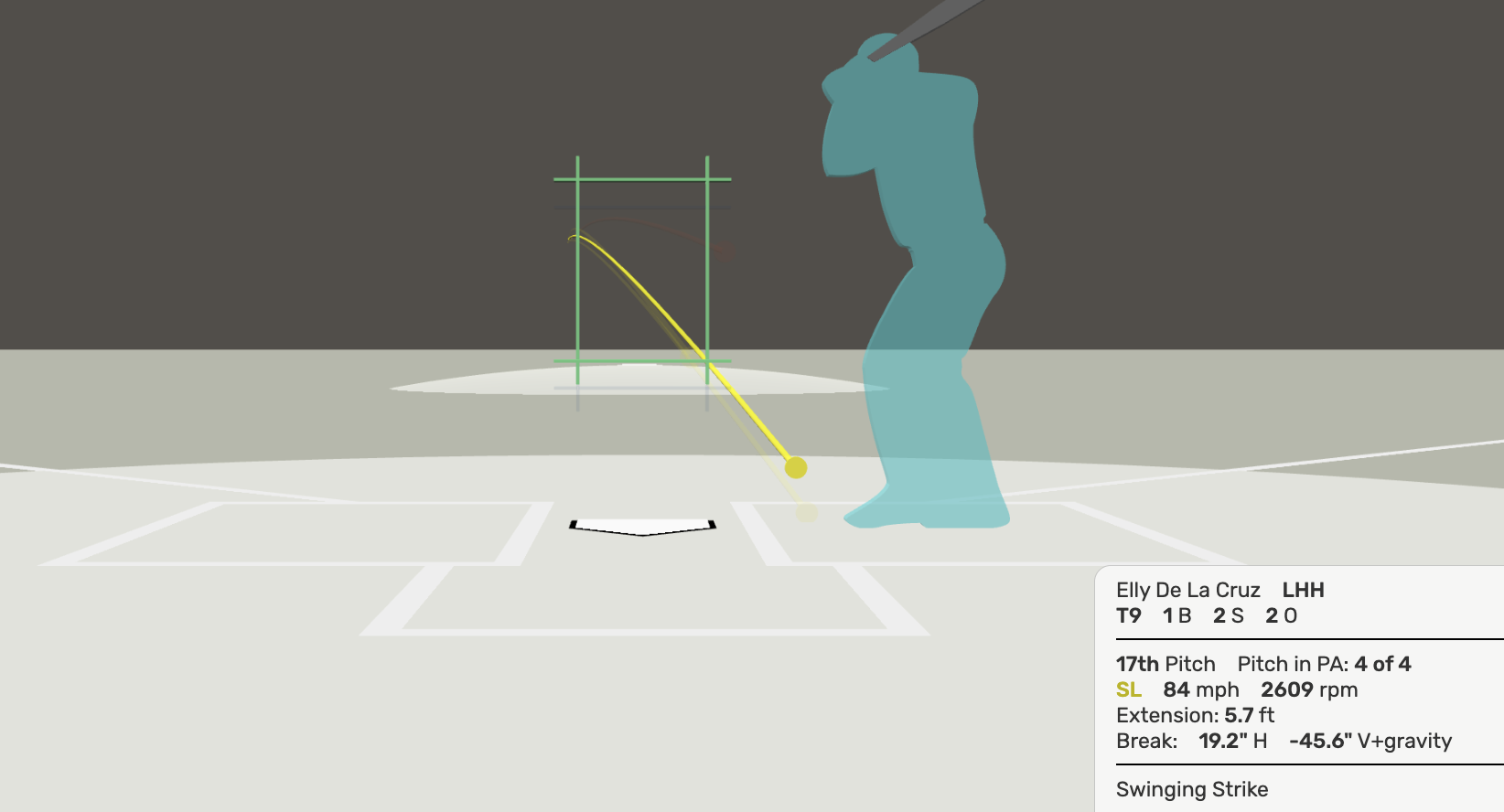}
    \caption{MLB Trackman. sites:https://reurl.cc/WxOWxZ}
    \label{fig:image1}
\end{figure}

\begin{figure}[htbp]
    \centering
    \includegraphics[width=0.4\textwidth]{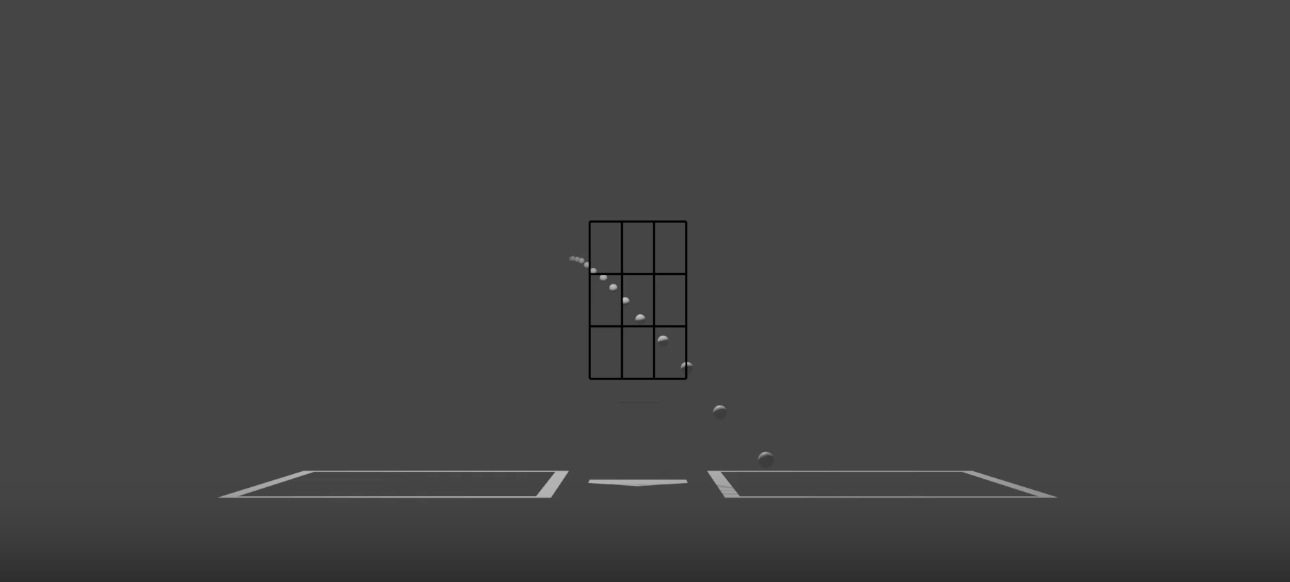}
    \caption{My results}
    \label{fig:image2}
\end{figure}

\section{Conclusion}
In conclusion, this study presents a novel neural network-based approach for tracking and reconstructing baseball pitch trajectories from 2D video footage to 3D coordinates. Our approach demonstrates promising results, achieving accurate reconstruction of pitch trajectories with a single 2D video input. By utilizing timestamp information and pixel-level tracking techniques, we compensate for the lack of depth information inherent in single-view setups.

Our method offers several advantages over traditional approaches, including simplicity, efficiency, and cost-effectiveness. By eliminating the need for multiple high-speed cameras and complex calibration techniques, our approach provides a more accessible alternative for tracking and analyzing baseball pitch trajectories, particularly in resource-limited environments.

However, it is important to acknowledge the limitations of our approach. The accuracy of reconstructed trajectories may still be affected by errors introduced by timestamp information and pixel-level tracking. Future research could explore methods to mitigate these limitations and further improve the accuracy and robustness of our approach.

Additionally, our framework can be extended to track and reconstruct trajectories of other sports-related objects, such as soccer balls or basketballs. Further investigation into the applicability of our approach to different sports domains could provide valuable insights and expand the scope of its potential applications.

Overall, this study contributes to the field of sports analytics by offering a promising avenue for simplifying and enhancing the analysis of baseball pitch trajectories. Further research and development in this area could lead to significant advancements in sports tracking and analytics.


\end{document}